\pdfoutput=1

\documentclass[11pt]{article}

\usepackage[final]{ACL2023}
 
\usepackage{microtype}
\usepackage{hyperref}
\usepackage{nameref}

\usepackage{times}
\usepackage{latexsym}
\usepackage{graphicx}
\usepackage{algorithm, algorithmic}
\usepackage{amsmath}
\usepackage{booktabs}
\usepackage{tabularx}
\usepackage{colortbl}
\usepackage{hhline}
\usepackage{amssymb} 
\usepackage{array} 
\usepackage{multirow} 
\usepackage{enumitem}

\usepackage[T1]{fontenc}
\usepackage[most]{tcolorbox}
\newtcolorbox{prompt}[1]{
    enhanced,
    drop shadow=black!5!white,
    left=4mm,
    right=4mm,
    top=2mm,
    bottom=2mm,
    boxsep=0mm,
    rounded corners,
    title=#1,
    fontupper=\footnotesize\linespread{0.9}\fontfamily{lmr}\selectfont,
    }

\usepackage[utf8]{inputenc}

\usepackage{inconsolata}

\title{CamPilot: A Multi-Agent Cinematic Assistant for Camera-Controlled Movie Generation}

\author{
\textbf{Yang Wu}\textsuperscript{$\clubsuit$}\thanks{\, Initial work was conducted while Y.W.\ was an intern at Adobe Research.} \quad
\textbf{Stefano Petrangeli}\textsuperscript{$\heartsuit$} \quad
\textbf{Ishita Dasgupta}\textsuperscript{$\heartsuit$}
\quad
\textbf{Yu Shen}\textsuperscript{$\heartsuit$}\thanks{\, Corresponding author.} \\
\textsuperscript{$\clubsuit$}Worcester Polytechnic Institute, Worcester, MA, USA \\
\textsuperscript{$\heartsuit$}Adobe Research, San Jose, CA, USA \\
\texttt{ywu19@wpi.edu}
\quad
\texttt{\{petrange, idasgupt, shenyu\}@adobe.com}
}

\begin{document}
\maketitle

\begin{abstract}
The integration of large language models (LLMs) into video generation has enabled rapid text-to-video creation and improved visual quality. However, it still falls short of professional filmmaking, where cinematographic language is less refined than human-crafted camera work and multi-shot continuity remains challenging. To address these limitations, we introduce CamPilot, a multi-agent framework that integrates cinematographic planning and camera-work control to produce more coherent, logically structured, and human-aesthetic movies. CamPilot adopts a GRPO-based learning paradigm to learn camera work planning from 14K real-world professional movies, internalizing motion patterns and composition principles that support reasoning over shooting techniques (e.g., camera angle, motion, and focal behavior) and cross-shot relationships for controllable camera-viewpoint generation. Multiple agents further collaborate and evolve to improve overall output quality. To support this work and further studies in this domain, we establish CamEval, a benchmark for evaluating camera work quality and cinematic engagement. Empirical results show that CamPilot outperforms state-of-the-art text-to-movie generation methods on cinematographic control and quality, highlighting the impact of professional camera design on movie generation.
\end{abstract}

\section{Introduction}
\label{sec:introduction}

\begin{figure}[t]
\centering 
\includegraphics[width=\linewidth]{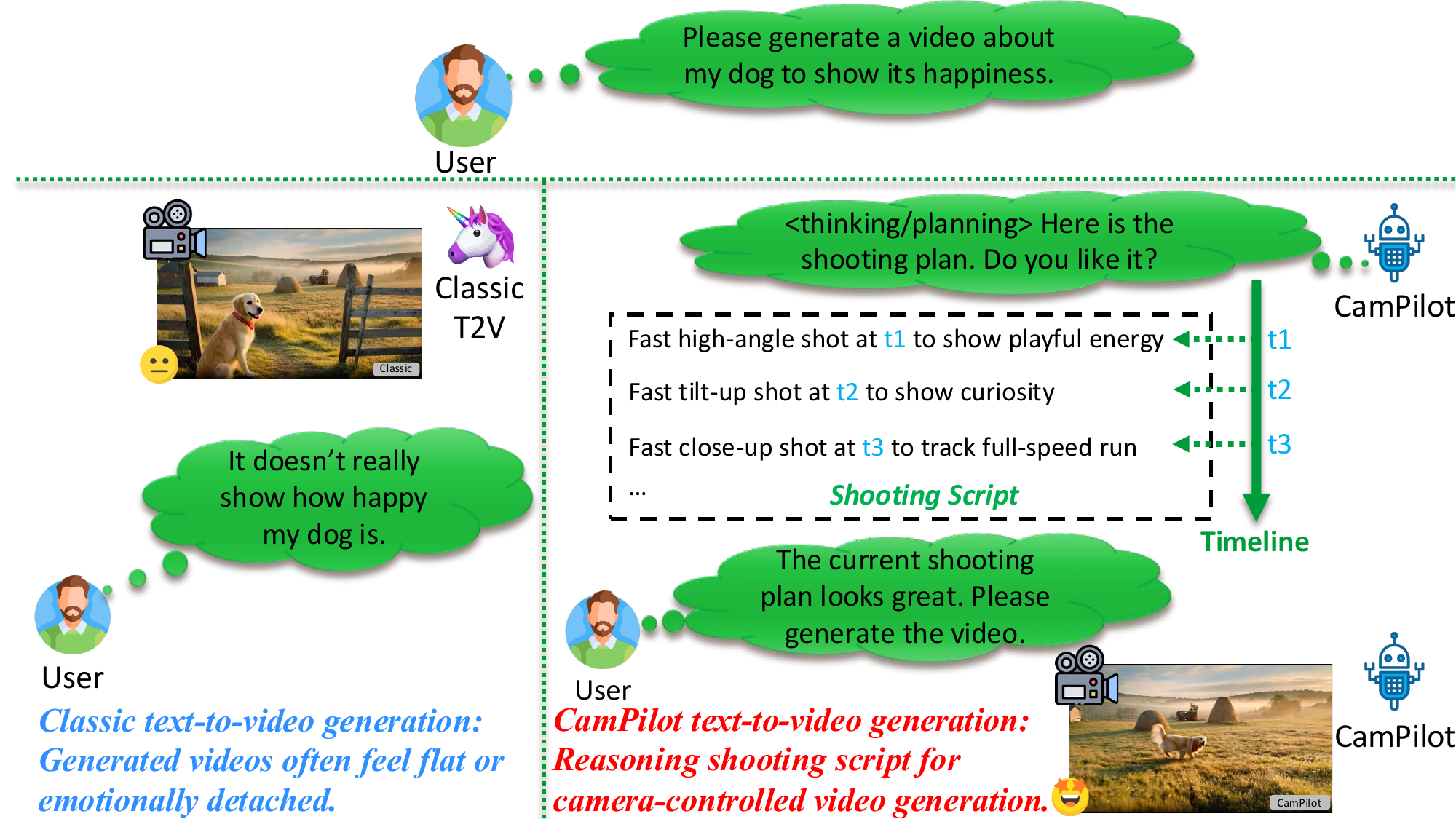}
\caption{Comparison of a classic text-to-video generation model and CamPilot for text-to-movie generation. The classic model (left) directly generates a video with generic, flat shots that often miss the user’s intent, whereas CamPilot (right) first performs reasoning to plan a professional shooting script and then generates a camera-controlled movie that better expresses the intended scene and emotion.}
\label{fig:intro}
\end{figure}

Large language models (LLMs) are increasingly integrated into text-conditioned video generation systems, enabling rapid text-to-video creation and improved visual quality~\citep{kondratyuk2023videopoet,bar2024lumiere}. These advances are reshaping creative workflows by allowing creators to iterate from a textual concept to a rendered clip with reduced manual effort~\citep{bar2024lumiere,xie2024dreamfactory}. They are particularly impactful for movie-like content creation, where users often expect not only photorealistic frames but also expressive camera work that conveys intent and emotion~\citep{wu2025automated,he2024cameractrl}. However, existing methods still fall short of professional filmmaking, where cinematographic language remains less refined than human-crafted shooting and multi-shot continuity is particularly difficult to maintain~\citep{xie2024dreamfactory,wu2025automated,li2024can}. This limitation reflects a mismatch between how current text-to-video models generate content and how professional filmmakers design camera work for coherent movies~\citep{he2024cameractrl,kuang2024collaborative,li2024can}. Once the story intent is known, human filmmakers typically prepare in advance by drafting a shooting script or shot list that specifies camera angle, motion, and focal behavior for each shot and coordinates how shots connect across a sequence~\citep{courant2024exceptional,xing2025motioncanvas}. We argue that text-to-movie generation should follow a similar planning-first principle, achieving stronger cinematographic expressiveness while preserving coherent multi-shot structure~\citep{lin2023videodirectorgpt,xie2024dreamfactory,wu2025automated}.

Motivated by this insight, we introduce \textit{CamPilot}, a multi-agent framework for text-to-movie generation that integrates professional cinematographic planning and camera-work control to produce more coherent, logically structured, and human-aesthetic movies~\citep{hu2024storyagent}. As illustrated in Figure~\ref{fig:intro}, a classic text-to-video generation model directly produces a video with generic, flat shots, which can miss the user’s intent even when the visual content is plausible. In contrast, CamPilot first performs reasoning to plan a professional shooting script, including shot-specific choices of camera angle, motion, and focal behavior, and then generates a camera-controlled movie that better expresses the intended scene and emotion~\citep{geng2025motion,feng2024i2vcontrol}. CamPilot adopts a GRPO-based~\citep{shao2024deepseekmath} learning paradigm to learn camera grammar from 14K real-world professional movies~\citep{bain2020condensed}, internalizing motion patterns and composition principles that support reasoning over shooting techniques and cross-shot relationships for controllable camera-viewpoint generation. Multiple agents further collaborate and evolve to improve overall output quality. For example, given a prompt that requires conveying a clear emotional intent, CamPilot can plan a time-ordered sequence with complementary shot types and transitions, then execute the plan under explicit camera control to maintain continuity and strengthen expression~\citep{ling2025vmbench,chen2025t2vworldbench}.

To support this work and further studies in this domain, we establish \textit{CamEval}, a benchmark for evaluating camera work quality and cinematic engagement in text-to-movie generation. CamEval is designed to assess both shot-level camera execution and multi-shot consistency, aligning evaluation with the challenges of cinematographic language and continuity~\citep{babu2025dynamiceval}. Using CamEval, we conduct extensive experiments against state-of-the-art text-to-movie generation methods. The results demonstrate that CamPilot improves cinematographic control and overall quality over strong baselines, highlighting the effectiveness of professional camera design for movie generation.

Our contributions in this paper are threefold and can be summarized as follows:

$\bullet$ We propose CamPilot, a multi-agent text-to-movie framework that unifies cinematographic planning with explicit camera-work control. It learns camera grammar from 14k real-world professional movies via a GRPO-based training paradigm, enabling shot-level technique reasoning and cross-shot relationship modeling for controllable camera-viewpoint generation.  \par
$\bullet$ We introduce CamEval, a benchmark for evaluating camera work quality and cinematic engagement, and we use it to compare CamPilot with state-of-the-art text-to-movie generation methods. \par
$\bullet$ Empirical experiments validate that camPilot not only improves professional camera-work quality, but also yields strong video generation quality, enhancing overall visual fidelity and multi-shot coherence. \par

\section{Related Work}

\subsection{Video Generation}
Recent progress in text-to-video generation is largely driven by diffusion-based models, which extend image diffusion to the spatiotemporal setting and improve realism and temporal coherence.
Early work such as Video Diffusion Models \citep{ho2022video} and Make-A-Video \citep{singer2022makeavideo} established foundational architectures for modeling motion and appearance, while latent-space formulations improve efficiency and scalability \citep{blattmann2023align}.
Building on large-scale text-to-image pretraining, recent systems further enhance quality via stronger temporal modules and data scaling, including ModelScopeT2V \citep{wang2023modelscope}, LaVie \citep{wang2023lavie}, and Stable Video Diffusion \citep{blattmann2023svd}. In parallel, transformer-style video language models tokenize visual content and learn long-range dependencies with autoregressive or decoder-only objectives, enabling flexible conditioning and multimodal generation \citep{yan2021videogpt,kondratyuk2023videopoet}.
Despite these advances, most text-to-video backbones focus on short clips and prioritize visual fidelity over cinematographic intent, leaving professional camera work and multi-shot continuity under-specified at generation time.
This motivates frameworks that explicitly model camera language and cross-shot structure on top of strong video generators.

\subsection{LLM-based Agents}
Large language models have become increasingly effective as agentic planners that decompose tasks, reason over intermediate states, and invoke tools \citep{brown2020gpt3,wei2022cot}.
Tool-augmented paradigms improve grounded decision-making by integrating external actions, including learning to call tools from supervision or self-generated traces \citep{nakano2021webgpt,schick2023toolformer,shen2023hugginggpt,qin2023toollm}.
To improve robustness, iterative refinement and self-critique loops have been explored for language agents, where agents revise outputs based on feedback or reflections \citep{shinn2023reflexion,yao2023tot}.
Beyond single-agent setups, multi-agent systems coordinate specialized roles to solve complex tasks through communication and division of labor \citep{park2023generativeagents,hong2023metagpt,wang2023voyager,li2023camel,qian2023chatdev}.
These agentic abstractions are particularly relevant to long-horizon generation problems, where planning, verification, and revision naturally map to distinct roles in a production pipeline.

\subsection{LLMs for Movie Generation}
Motivated by the need for long-form and multi-scene consistency, recent work uses LLMs to expand a user prompt into structured scripts, storyboards, or scene plans, and then conditions downstream generators on these intermediate representations.
VideoDirectorGPT \citep{lin2023videodirectorgpt} demonstrates LLM-guided multi-scene planning with explicit layouts, and VideoStudio \citep{long2024videostudio} similarly leverages LLMs to produce multi-scene scripts to improve content consistency.
Several multi-agent pipelines further decompose the process into specialized roles and incorporate iterative refinement to improve long-video coherence, including DreamFactory \citep{xie2024dreamfactory}, Mora \citep{mora2024}, and StoryAgent \citep{hu2024storyagent}.
More explicitly film-oriented systems simulate filmmaking roles such as director, screenwriter, and cinematographer, for example MovieAgent \citep{wu2025automated} and FilmAgent \citep{xu2025filmagent}, while Anim-Director targets controllable animation production via an agentic workflow \citep{li2024animdirector}. Additionally, spatio-temporal event modeling frameworks \citep{cen2025pde} demonstrate the benefit of hierarchical reasoning over complex dynamic processes. While these systems improve narrative structure and multi-scene consistency, they typically treat camera work as a prompted attribute or a heuristic control signal, rather than learning camera grammar from real professional footage with a trainable objective.
Our work complements this line by focusing on learning camera language from real movies and using it to support controllable, cross-shot cinematographic planning within a multi-agent text-to-movie pipeline.

\section{Methodology}
\label{sec:methodology}

\subsection{Problem Definition}
Given a textual prompt $S$, the goal of camera-controlled long-form movie generation is to produce a multi-scene, multi-shot movie $\hat{V}$ that is coherent in narrative and cinematic style.
We aim to learn a mapping function
\begin{equation}
    F: S \rightarrow \hat{V}.
\end{equation}
The output $\hat{V}$ is a sequence of shot clips organized by scenes:
\begin{equation}
    \hat{V} = \left\{ \hat{v}^{\,i}_{j} \;\middle|\; i=1,\ldots,N,\; j=1,\ldots,M_i \right\},
\end{equation}
where $\hat{v}^{\,i}_{j}$ denotes the $j$-th shot video in the $i$-th scene, $N$ is the number of scenes, and $M_i$ is the number of shots in scene $i$.
Function $F(\cdot)$ instantiates a hierarchical pipeline that (i) expands $S$ into a structured story with scene and shot plans, and (ii) generates each shot under explicit camera-work control, so that all shots can be concatenated into the final movie.

\subsection{CamPilot Overview}
\label{sec:campilot_overview}
\begin{figure*}[t]
  \centering
  \includegraphics[width=\textwidth]{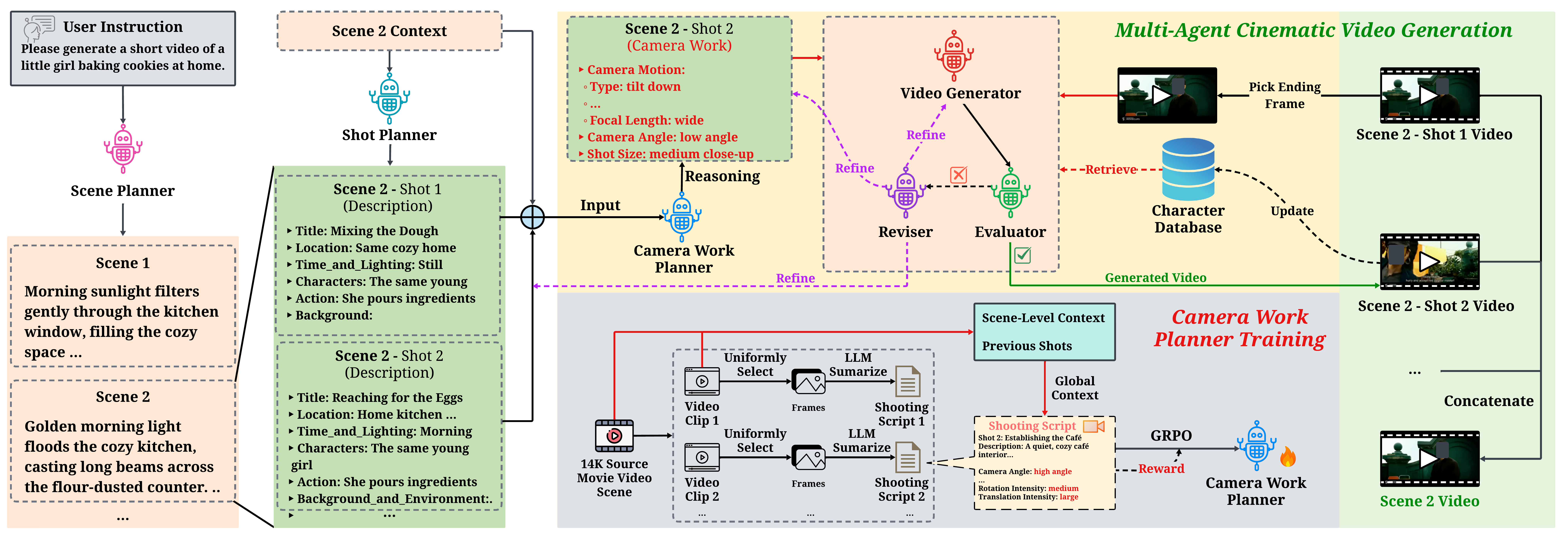}
  \caption{Overall framework of CamPilot. Given a user prompt, CamPilot first plans a scene-level storyline and a shot-level script. A trainable camera work planner then performs reasoning over the current shot and its context to produce structured camera work, which conditions a backbone video generator. An evaluator--reviser loop iteratively refines the shot description, camera work, or generation conditions until quality checks are satisfied. A character bank and frame-to-frame conditioning support character consistency and cross-shot continuity for long-form movie generation.}
  \label{fig:framework}
\end{figure*}

We present \textit{CamPilot}, a multi-agent framework for text-to-movie generation, as illustrated in Figure~\ref{fig:framework}.
The key idea is to separate \emph{story planning} from \emph{cinematographic execution}.
CamPilot first expands the user prompt into a scene-level storyline and a shot-level script, then invokes a trainable \emph{Camera Work Planner} to perform reasoning over the current shot and its context to produce structured camera work.
A video generator (any off-the-shelf text-to-video backbone) then synthesizes the shot conditioned on the planned camera work.
Finally, an evaluator--reviser loop iteratively refines the shot description, camera work, or generation prompt until the generated shot passes quality checks.
A character bank and frame-to-frame conditioning are used to improve character consistency and cross-shot continuity for long-form movie synthesis.

\subsubsection{Scene Planner}
\label{sec:scene_planner}
The \textbf{Scene Planner} expands the user prompt $S$ into a scene outline
\begin{equation}
\mathcal{C} = \{c_1,\ldots,c_N\} = \Pi_{\mathrm{scene}}(S),
\end{equation}
where each $c_i$ is a scene-level description that specifies the high-level narrative intent, location, time, and overall mood of the scene.
This stage provides a global scaffold that constrains later shot-level planning and helps maintain narrative coherence across scenes.

\subsubsection{Shot Planner}
\label{sec:shot_planner}
Given a scene description $c_i$, the \textbf{Shot Planner} decomposes the scene into an ordered shot list:
\begin{equation}
\mathcal{D}_i = \{d_{i,1},\ldots,d_{i,M_i}\} = \Pi_{\mathrm{shot}}(c_i, S),
\end{equation}
where each shot description $d_{i,j}$ is a structured script containing fine-grained elements, including
Title, Location, Time and Lighting, Characters, Action,
Background and Environment, Mood and Tone, and Intention.
This representation makes the shot intent explicit and provides sufficient context for downstream camera-work reasoning.

\subsubsection{Camera Work Planner}
\label{sec:camera_planner}
The \textbf{Camera Work Planner} is the core trainable agent in CamPilot.
For the $j$-th shot in scene $i$, we construct its planning context from three sources:
(i) the scene-level description $c_i$,
(ii) the preceding shots in the same scene $\mathcal{D}_{i,<j}=\{d_{i,1},\ldots,d_{i,j-1}\}$,
and (iii) the current shot description $d_{i,j}$.
We denote the input context as
\begin{equation}
x_{i,j} = (c_i,\; \mathcal{D}_{i,<j},\; d_{i,j}).
\label{eq:cam_context}
\end{equation}
The planner then outputs structured camera work
\begin{equation}
w_{i,j} = \Pi_{\mathrm{cam}}(x_{i,j}),
\end{equation}
where $w_{i,j}$ consists of three main components:
\begin{equation}
w_{i,j} = (a_{i,j},\; s_{i,j},\; m_{i,j}),
\end{equation}
with camera angle $a_{i,j}$, shot size $s_{i,j}$, and camera motion $m_{i,j}$.
We further represent motion as a 5-tuple
\begin{equation}
m_{i,j} = (\tau_{i,j},\; f_{i,j},\; u_{i,j},\; \rho_{i,j},\; \delta_{i,j}),
\end{equation}
corresponding to motion type $\tau$, focal length $f$, motion speed $u$, rotation intensity $\rho$, and translation intensity $\delta$.
In Section~\ref{sec:dataset_construction}, we describe how we construct supervision from real-world movies and train $\Pi_{\mathrm{cam}}$ with a GRPO-based learning paradigm using a structured reward that factorizes over these attributes.
\subsubsection{Training the Camera Work Planner with GRPO}
\label{sec:grpo_training}
Reinforcement learning has shown strong potential in optimizing specialized LLM policies~\citep{wu2024knowledge,yao2025elevating}. We train the Camera Work Planner with a GRPO-based learning paradigm, which optimizes a policy to maximize task-specific rewards under a stable, regularized update.
Let $\pi_\theta$ denote the planner policy with parameters $\theta$, which maps the planning context $x_{i,j}$ (Equation~\ref{eq:cam_context}) to a camera-work output $w_{i,j}$.
We treat each camera-work attribute as a discrete decision and define the training objective over samples drawn from the current policy:
\begin{equation}
\begin{aligned}
\mathcal{L}_{\mathrm{GRPO}}(\theta)
&=\;
-\mathbb{E}_{w \sim \pi_\theta(\cdot \mid x)}
\big[ R(x,w) \big] \\
&\quad+\;
\beta \, \mathrm{KL}\!\left(\pi_\theta(\cdot \mid x)\;\|\;\pi_{\mathrm{ref}}(\cdot \mid x)\right),
\end{aligned}
\label{eq:grpo_loss}
\end{equation}
where $x$ denotes a planner input context, $R(x,w)$ is the reward for predicting camera work $w$ under context $x$, $\pi_{\mathrm{ref}}$ is a fixed reference policy, $\mathrm{KL}(\cdot\|\cdot)$ is the Kullback--Leibler divergence, and $\beta$ controls the strength of regularization.

\paragraph{Reward design.}
Given the ground-truth camera work label
$
w^\star = (a^\star, s^\star, m^\star)
$
from our dataset (Section~\ref{sec:dataset_construction}) and the planner prediction
$
w = (a, s, m),
$
we compute a structured reward that factorizes over camera-work attributes:
\begin{equation}
\begin{aligned}
R(x,w)
&=\;
\frac{1}{3} r_{\mathrm{angle}}(a, a^\star)
\;+\;
\frac{1}{3} r_{\mathrm{size}}(s, s^\star) \\
&\quad+\;
\frac{1}{3} r_{\mathrm{motion}}(m, m^\star).
\end{aligned}
\label{eq:reward_main}
\end{equation}
Here $r_{\mathrm{angle}}$ and $r_{\mathrm{size}}$ measure whether the predicted camera angle and shot size match the corresponding labels.
For camera motion, we use the same factorization as our output space, where
$
m=(\tau,f,u,\rho,\delta)
$
and
$
m^\star=(\tau^\star,f^\star,u^\star,\rho^\star,\delta^\star)
$
denote the predicted and labeled motion attributes, respectively.
We define the motion reward as an average over the five motion attributes:
\begin{equation}
\begin{aligned}
r_{\mathrm{motion}}(m, m^\star)
&=\;
\frac{1}{5} r_{\mathrm{type}}(\tau,\tau^\star)
+\frac{1}{5} r_{\mathrm{focal}}(f,f^\star) \\
&+\;
\frac{1}{5} r_{\mathrm{speed}}(u,u^\star)
+\frac{1}{5} r_{\mathrm{rot}}(\rho,\rho^\star) \\
&+\;
\frac{1}{5} r_{\mathrm{trans}}(\delta,\delta^\star).
\end{aligned}
\label{eq:reward_motion}
\end{equation}
Each component reward $r(\cdot,\cdot)$ compares the predicted class against the labeled class for the corresponding attribute.
This design encourages the planner to learn camera grammar at both the coarse level (angle, size, motion) and the fine-grained level (motion sub-attributes), while producing a complete, structured camera-work specification.

\subsubsection{Agent Evolution via Evaluation and Revision}
\label{sec:evolution}
Long-form generation is sensitive to accumulated errors, such as inconsistent characters, abrupt camera changes, or visually unstable motion.
CamPilot addresses this with an \textbf{evolution loop} that iteratively evaluates and revises each shot until it meets quality requirements.

\paragraph{Evaluator.}
After generating $\hat{v}^{\,i}_{j}$, an evaluator $E$ inspects the shot and returns a binary decision and textual feedback:
\begin{equation}
(\mathrm{ok}_{i,j},\; g_{i,j}) = E(\hat{v}^{\,i}_{j}),
\label{eq:evaluator}
\end{equation}
where $\mathrm{ok}_{i,j}\in\{0,1\}$ indicates whether the shot passes checks, and $g_{i,j}$ describes failure reasons (e.g., inconsistency or lack of smoothness).

\paragraph{Reviser.}
If $\mathrm{ok}_{i,j}=0$, a reviser agent $R$ takes the feedback $g_{i,j}$ and produces a refinement by editing one of three targets:
the shot description $d_{i,j}$, the planned camera work $w_{i,j}$, or the generator-side prompt/conditions used by $G$:
\begin{equation}
(d_{i,j}', w_{i,j}', \text{cond}') = R(d_{i,j}, w_{i,j}, g_{i,j}).
\label{eq:reviser}
\end{equation}
We then re-generate the shot with the refined inputs and repeat the evaluate--revise cycle up to a maximum number of iterations.
The accepted shot is appended to the movie sequence, and its ending frame is used to condition the next shot, enabling long-form synthesis with improved cinematic continuity.

\subsubsection{Character Bank and Video Generation}
\label{sec:gen}
To improve identity consistency, inspired by memory modeling in specialized LLM agents~\citep{wu2025teaching}, CamPilot maintains a \textbf{Character Bank} $\mathcal{B}$ that stores a reference item for each character, such as a character identifier and a reference image.
For shot $(i,j)$, we retrieve relevant character references $\mathcal{B}_{i,j}$ based on $d_{i,j}$ and incorporate them into generation conditions.
We then construct the generation input for a backbone video generator $G$ as
\begin{equation}
\hat{v}^{\,i}_{j} = G\big(d_{i,j},\; w_{i,j},\; \mathcal{B}_{i,j},\; \hat{e}^{\,i}_{j-1}\big),
\label{eq:video_gen}
\end{equation}
where $\hat{e}^{\,i}_{j-1}$ denotes the ending frame of the previous shot in the same scene.
We use $\hat{e}^{\,i}_{j-1}$ as the start-frame condition for the current shot whenever available, which encourages smoother transitions and better cross-shot continuity for long-form movie generation.
Backbone $G$ can be instantiated by any existing text-to-video model, and CamPilot focuses on improving controllability and cinematic structure through planning and refinement.

\section{Dataset Construction}
\label{sec:dataset_construction}

Learning professional camera grammar requires scalable supervision that links real-world movie footage to structured camera-work attributes. To this end, we construct \textit{CamEval}, a camera-work dataset distilled from 14K real-world professional movies, where each instance pairs a clip-level visual observation with a structured shooting script and camera-work labels. Starting from the condensed movie collection, we first segment each source movie into a set of short clips. For each clip, we uniformly sample a sequence of frames to capture both appearance and motion cues while keeping annotation cost manageable. We then prompt a vision--language model, \texttt{Qwen/Qwen3-VL-32B-Instruct}~\citep{bai2025qwen3}, to summarize the sampled frames into a structured shooting script, including (i) a compact description of the scene and shot (e.g., location, characters, action, mood, and intention) and (ii) the corresponding camera work. The camera-work labels follow the same schema as our Camera Work Planner output space: camera angle, shot size, and camera motion, where motion is further decomposed into motion type, focal length, motion speed, rotation intensity, and translation intensity. These VLM-produced shooting scripts provide training supervision for our camera-work planner under the same contextual inputs used in CamPilot, including the scene-level description, preceding shots as cross-shot context, and the current shot description. The detailed data statistics are provided in Table~\ref{tab:data_stats}. We use these labels to train the Camera Work Planner with a GRPO-based objective (Section~\ref{sec:grpo_training}).

\begin{table}[t]
\centering

\setlength{\tabcolsep}{6pt}
\renewcommand{\arraystretch}{1.2}
\begin{tabular}{l r}
\hline
\# Videos & 14,581 \\
\# Shots & 99,975 \\
Movie years & 2019--2020 \\
Movie Genres & general \\
\hline
\# Train videos & 12,641 \\
\# Train shots & 85,048 \\
\# Validation videos & 1,000 \\
\# Validation shots & 6,663 \\
\# Test videos & 1,000 \\
\# Test shots & 6,578 \\
\hline
\end{tabular}
\caption{Data statistics of CamEval.}
\label{tab:data_stats}
\end{table}


\section{Experiments}
\label{sec:experiment}

\begin{table*}[t]
    \centering
    \renewcommand{\arraystretch}{1.2}
    \resizebox{1.0\textwidth}{!}{
    \begin{tabular}{@{}llcccccccc@{}}
        \toprule
        \multirow{2}{*}[-1ex]{\shortstack[l]{CameraWorkPlanner\\ LLM Backbone}} 
        & \multirow{2}{*}[-1ex]{Method} 
        & \multicolumn{4}{c}{Camera Work Classification (\%)~$\uparrow$} 
        & \multicolumn{2}{c}{Keyframe Generation (\%)~$\uparrow$} & \multicolumn{2}{c}{Movie Generation (\%)~$\uparrow$} \\
        \cmidrule(lr){3-6} \cmidrule(lr){7-8} \cmidrule(lr){9-10}
        & & Macro-Acc & Macro-Rec & Macro-Prec & Macro-F1 
          & CLIP & Inception & Sub\_Cons & Aesthetic \\
        \midrule
        \midrule
\multicolumn{10}{l}{\emph{Baselines with Closed-Source LLMs}}\\
        \midrule
           Deepseek-R1 & Standard            
           & 52.7 (0.1) & 40.5 (0.8) & 45.7 (2.0) & \underline{36.0 (0.8)}
           & 20.9 (0.4) & 9.4 (0.3) & 91.2 (0.5) & 55.8 (0.6) \\
           Claude-Sonnet-4.5 & Standard         
           & \textbf{60.1 (0.1)} & \textbf{51.6 (1.3)} & \textbf{55.0 (2.2)} & \textbf{44.8 (1.1)}
           & \underline{21.8 (0.4)} & \underline{9.9 (0.3)} & \underline{93.4 (0.4)} & \underline{57.9 (0.6)} \\
          GPT-5  & Standard              
          & \underline{53.2 (0.1)} & \underline{41.9 (1.7)} & \underline{47.0 (3.9)} & 35.4 (1.9)
          & \textbf{22.0 (0.4)} & \textbf{10.1 (0.3)} & \textbf{94.0 (0.4)} & \textbf{58.6 (0.6)} \\
        \midrule
        \midrule
\multicolumn{10}{l}{\emph{Baselines and CamPilot with the same Open-Source LLM Backbones}}\\
        \midrule
        \multirow{5}{*}{Llama-3.1-8B-Instruct} 
            & Standard         
            & 45.5 (0.0) & 12.2 (0.6) & 13.1 (0.1) & 9.8 (0.3)
            & 18.9 (0.5) & 8.6 (0.4) & 87.6 (0.6) & 51.2 (0.8) \\
            & Vanilla-SFT            
            & \underline{61.4 (0.2)} & 15.1 (1.2) & 14.9 (1.6) & 13.7 (1.1)
            & \underline{20.8 (0.4)} & \underline{9.3 (0.3)} & \underline{92.1 (0.5)} & \underline{56.0 (0.7)} \\
            & DreamFactory         
            & 58.0 (0.2) & 20.0 (1.3) & 19.0 (1.2) & 18.5 (1.1)
            & 20.2 (0.4) & 9.1 (0.3) & 91.0 (0.5) & 54.8 (0.7) \\
            & MovieAgent              
            & 59.5 (0.2) & \underline{22.0 (1.2)} & \underline{21.0 (1.1)} & \underline{20.0 (1.0)}
            & 20.6 (0.4) & 9.2 (0.3) & 91.4 (0.5) & 55.4 (0.7) \\
            & \textbf{CamPilot (Ours)} 
            & \textbf{63.7 (0.2)} & \textbf{28.5 (1.0)} & \textbf{27.1 (0.8)} & \textbf{27.2 (0.9)}
            & \textbf{22.0 (0.3)} & \textbf{9.9 (0.3)} & \textbf{95.0 (0.4)} & \textbf{59.0 (0.6)} \\
        \midrule
        \multirow{5}{*}{Gemma-3-12b-it} 
            & Standard         
            & 56.7 (0.2) & 38.8 (2.7) & 34.7 (2.5) & 34.2 (2.5)
            & 20.4 (0.5) & 9.4 (0.4) & 90.6 (0.6) & 54.7 (0.8) \\
            & Vanilla-SFT            
            & \underline{62.6 (0.0)} & 37.3 (1.9) & \underline{36.5 (1.2)} & \textbf{36.1 (1.4)}
            & \underline{21.2 (0.4)} & \underline{9.8 (0.3)} & \underline{93.0 (0.5)} & \underline{56.8 (0.7)} \\
            & DreamFactory         
            & 60.8 (0.1) & 35.2 (2.2) & 33.9 (2.0) & 33.0 (2.0)
            & 20.8 (0.4) & 9.6 (0.3) & 92.1 (0.5) & 55.8 (0.7) \\
            & MovieAgent             
            & 61.9 (0.1) & \underline{36.5 (2.1)} & \textbf{37.2 (1.9)} & \underline{34.5 (1.9)}
            & 21.0 (0.4) & 9.7 (0.3) & 92.5 (0.5) & 56.2 (0.7) \\
            & \textbf{CamPilot (Ours)} 
            & \textbf{66.0 (0.2)} & \textbf{40.3 (0.8)} & \textbf{38.6 (3.0)} & 33.6 (0.6)
            & \textbf{22.3 (0.3)} & \textbf{10.2 (0.3)} & \textbf{95.6 (0.4)} & \textbf{59.4 (0.6)} \\
        \midrule
        \multirow{5}{*}{Qwen2.5-7B-Instruct} 
            & Standard         
            & 46.4 (0.1) & 18.1 (0.1) & 19.8 (0.2) & 14.0 (0.1)
            & 19.2 (0.5) & 8.8 (0.4) & 88.4 (0.6) & 52.0 (0.8) \\
            & Vanilla-SFT            
            & \underline{60.9 (0.1)} & 23.3 (0.2) & 21.0 (0.2) & 20.6 (0.2)
            & \underline{20.9 (0.4)} & \underline{9.5 (0.3)} & \underline{92.4 (0.5)} & \underline{56.1 (0.7)} \\
            & DreamFactory        
            & 57.6 (0.1) & 26.0 (0.9) & 24.0 (0.8) & 23.5 (0.8)
            & 20.2 (0.4) & 9.2 (0.3) & 91.0 (0.5) & 54.3 (0.7) \\
            & MovieAgent             
            & 58.8 (0.1) & \underline{27.5 (0.8)} & \underline{25.6 (0.8)} & \underline{25.0 (0.8)}
            & 20.5 (0.4) & 9.3 (0.3) & 91.6 (0.5) & 55.0 (0.7) \\
            & \textbf{CamPilot (Ours)} 
            & \textbf{62.7 (0.0)} & \textbf{30.7 (0.3)} & \textbf{29.8 (4.0)} & \textbf{29.6 (0.4)}
            & \textbf{21.7 (0.3)} & \textbf{10.0 (0.3)} & \textbf{95.1 (0.4)} & \textbf{58.9 (0.6)} \\
        \bottomrule
    \end{tabular}
    }
    \caption{Comparative results with Automatic metrics on camera work classification and keyframe/movie generation. The mean (std) over three random runs is reported. The best results are shown in \textbf{bold}, and the second-best are \underline{underlined}.}
    \label{tab:main_results}
\end{table*}

\subsection{Experimental Settings}
Our CamPilot text-to-movie pipeline contains a scene planner, a shot planner, a trainable Camera Work Planner, and an evaluator--reviser loop for iterative refinement. Among these components, only the Camera Work Planner is trained, while the other agents operate via prompting and tool orchestration. The Camera Work Planner is trained on CamEval with GRPO using TRL, where we optimize the planner policy to maximize the structured camera-work reward defined in Section~\ref{sec:grpo_training}. We use AdamW with learning rate $2 \times 10^{-5}$ and batch size 8, and train for 4 epochs. Unless otherwise specified, we set temperature to 0.8 and top\_p to 1.0 during generation. Our default backbone video generator is Adobe Firefly, and we run three random seeds and report mean and standard deviation. Experiments are implemented using TRL~\citep{vonwerra2022trl}, Transformers~\citep{wolf2020transformers}, and PyTorch~\citep{paszke2019pytorch} on a 64-core CPU and eight 80GB A100 GPUs.

\subsection{Baselines}
We compare CamPilot with four representative baselines: Standard, Vanilla-SFT, DreamFactory~\citep{xie2024dreamfactory}, and MovieAgent~\citep{wu2025automated} (Table~\ref{tab:main_results}). Standard predicts structured camera work from the same planner input context without training. Vanilla-SFT fine-tunes the Camera Work Planner with supervised learning on CamEval to predict camera-work labels. DreamFactory is an in-context learning baseline that uses demonstrations to prompt the planner backbone to output camera work. MovieAgent is a chain-of-thought baseline that performs multi-step reasoning for camera planning. We conduct experiments on three open-source planner backbones: Llama-3.1-8B-Instruct~\citep{llama31_8b_instruct_modelcard}, Gemma-3-12B-it~\citep{gemma3_12b_it_modelcard}, and Qwen2.5-7B-Instruct~\citep{hui2024qwen2}. We also report Standard results with closed-source planner backbones, including DeepSeek-R1~\citep{deepseek_r1_modelcard}, Claude Sonnet 4.5~\citep{claude_sonnet_45_official}, and GPT-5~\citep{gpt5_official}. For all methods, we use the same planner input context (Equation~\ref{eq:cam_context}) and the same camera-work label schema (Section~\ref{sec:camera_planner}). Unless otherwise specified, we fix the downstream video generator to the same Firefly setting to isolate the effect of camera-work planning.

\subsection{Tasks and Metrics}
We evaluate all methods on three tasks with automatic metrics (Table~\ref{tab:main_results}). 
\textbf{Camera Work Classification} We formulate camera-work planning as a multi-attribute classification problem over camera angle, shot size, and camera motion. Camera motion is further factorized into motion type, focal length, motion speed, rotation intensity, and translation intensity. We report Macro-Acc, Macro-Rec, Macro-Prec, and Macro-F1, computed over the discrete label space defined by our camera-work schema.
\textbf{Keyframe Generation} We evaluate the visual quality of keyframes using CLIPScore based on CLIP~\citep{radford2021learning} and Inception Score~\citep{salimans2016improved}. 
\textbf{Movie Generation} We evaluate long-form generation quality using VBench~\citep{huang2024vbench}. Specifically, we report Subject Consistency (Sub\_Cons), which measures subject identity stability across consecutive shots, and Aesthetic, which evaluates the overall aesthetic quality of temporal transitions between shots.

\begin{figure*}[ht]
  \centering
  \includegraphics[width=\textwidth]{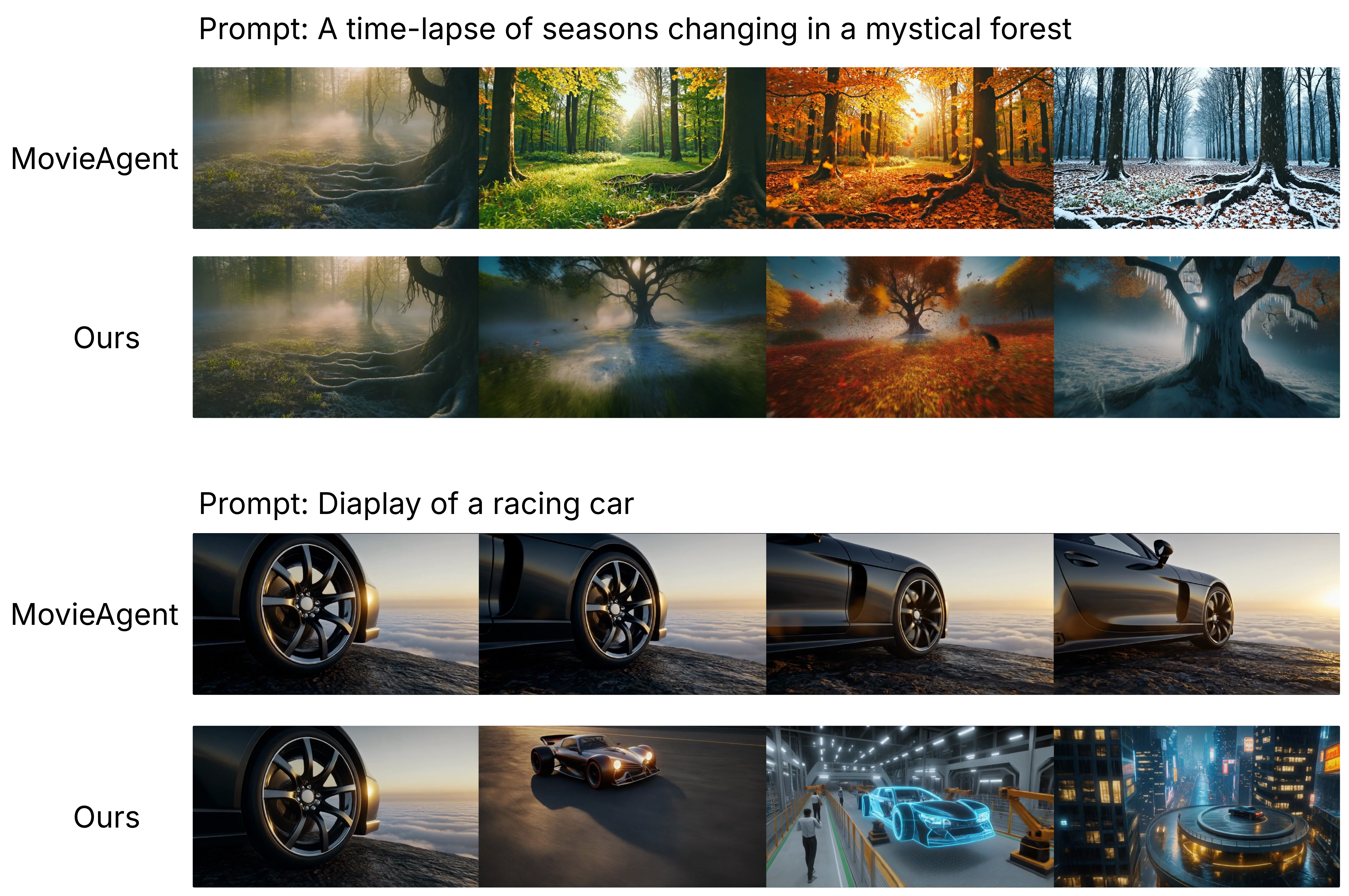}
  \caption{Qualitative comparison between MovieAgent and CamPilot.}
  \label{fig:qua}
\end{figure*}

\begin{table*}[t]
    \centering
    \renewcommand{\arraystretch}{1.2}
    \resizebox{1.0\textwidth}{!}{
    \begin{tabular}{@{}llcccccccc@{}}
        \toprule
        \multirow{2}{*}[-1ex]{LLM Backbone} 
        & \multirow{2}{*}[-1ex]{Method} 
        & \multicolumn{4}{c}{Camera Work Classification (\%)~$\uparrow$} 
        & \multicolumn{2}{c}{Keyframe Generation (\%)~$\uparrow$} & \multicolumn{2}{c}{Movie Generation (\%)~$\uparrow$} \\
        \cmidrule(lr){3-6} \cmidrule(lr){7-8} \cmidrule(lr){9-10}
        & & Macro-Acc & Macro-Rec & Macro-Prec & Macro-F1 
          & CLIP & Inception & Sub\_Cons & Aesthetic \\
        \midrule
        \midrule
        \multirow{5}{*}{Llama-3.1-8B-Instruct} 
            & Standard         
            & 44.3 (0.1) & 11.6 (0.6) & 12.5 (0.2) & 9.1 (0.3)
            & 18.9 (0.5) & 8.6 (0.4) & 87.6 (0.6) & 51.2 (0.8) \\
            & Vanilla-SFT            
            & \underline{59.8 (0.2)} & \underline{26.7 (2.0)} & \underline{25.4 (1.8)} & \underline{25.7 (1.9)}
            & \underline{20.8 (0.4)} & \underline{9.3 (0.3)} & \underline{92.1 (0.5)} & \underline{56.0 (0.7)} \\
            & DreamFactory         
            & 56.8 (0.2) & 19.1 (1.3) & 18.2 (1.2) & 17.8 (1.1)
            & 20.2 (0.4) & 9.1 (0.3) & 91.0 (0.5) & 54.8 (0.7) \\
            & MovieAgent              
            & 58.2 (0.2) & 21.1 (1.2) & 20.0 (1.1) & 19.2 (1.0)
            & 20.6 (0.4) & 9.2 (0.3) & 91.4 (0.5) & 55.4 (0.7) \\
            & \textbf{CamPilot (Ours)} 
            & \textbf{62.1 (0.2)} & \textbf{27.6 (1.1)} & \textbf{26.3 (0.8)} & \textbf{26.5 (0.9)}
            & \textbf{22.0 (0.3)} & \textbf{9.9 (0.3)} & \textbf{95.0 (0.4)} & \textbf{59.0 (0.6)} \\
        \bottomrule
    \end{tabular}
    }
    \caption{Single-shot results with Automatic metrics on camera work classification and keyframe/movie generation. The mean (std) over three random runs is reported. The best results are shown in \textbf{bold}, and the second-best are \underline{underlined}.}
    \label{tab:ablation_study_single_shot}
\end{table*}

\subsection{Experimental Results}
Table~\ref{tab:main_results} demonstrates that CamPilot consistently achieves the best camera-work classification performance when controlling for the same open-source backbone, indicating that the gains come from our training and planning framework rather than model size. For Llama-3.1-8B-Instruct, CamPilot improves Macro-Acc from 45.5 (Standard) to 63.7, and also yields clear gains over prompting baselines such as DreamFactory (58.0) and MovieAgent (59.5). Beyond classification, CamPilot also delivers the strongest downstream generation quality within each backbone family, achieving the best CLIP and Inception scores and improving cross-shot subject stability and aesthetics, as reflected by Sub\_Cons around 95 and Aesthetic around 59 across backbones (e.g., Llama: 95.0/59.0, Gemma: 95.6/59.4, Qwen: 95.1/58.9), which is consistent with our goal of learning camera grammar for coherent multi-shot planning. Importantly, CamPilot remains competitive against closed-source commercial LLMs despite the substantial parameter gap. In terms of Macro-Acc, CamPilot exceeds the strongest commercial baseline in our comparison, achieving 63.7 versus 60.1 from Claude-Sonnet-4.5, while also outperforming GPT-5 (53.2) and Deepseek-R1 (52.7). Although commercial models can be stronger on certain precision or recall dimensions, these differences are expected given their much larger capacity. Overall, the results confirm that CamPilot provides consistent and reliable accuracy improvements under fixed open-source backbones, while narrowing the performance gap to proprietary systems.

\subsection{Qualitative Results}

Figure~\ref{fig:qua} presents qualitative comparisons between CamPilot and baseline methods. Overall, CamPilot produces more dynamic and expressive camera work, with clearer motion intent and more coherent shot progression across scenes. Compared to baselines that often generate static or weakly varied camera behaviors, our results exhibit richer camera movements and more deliberate changes in shot composition, leading to videos that appear more cinematic and engaging. These qualitative examples align with the quantitative gains in camera work classification and generation metrics, and further demonstrate CamPilot’s ability to translate structured camera planning into visually compelling multi-shot videos.


\subsection{Ablation Study}
Table~\ref{tab:ablation_study_single_shot} shows that CamPilot remains best in the single-shot setting, improving Macro-Acc from 44.3 to 62.1 and Macro-F1 from 9.1 to 26.5, and outperforming Vanilla-SFT (59.8/25.7). It also achieves the strongest generation quality (CLIP/Inception 22.0/9.9) and higher Sub\_Cons/Aesthetic (95.0/59.0) than Standard (87.6/51.2).

\subsection{Discussion}

CamPilot frames camera-work planning as a context-aware, structured decision process. By jointly considering scene-level intent, preceding-shot context, and the current shot description, the Camera Work Planner produces camera specifications that are aligned not only with individual shot content but also with the progression of a multi-shot movie. The GRPO objective further supports this formulation by optimizing a factorized reward over complementary camera attributes, including angle, shot size, motion, focal behavior, speed, and motion intensity. This design is particularly beneficial for long-form generation, where effective cinematography requires coordinated transitions and shot relationships instead of independently selected camera tags.

CamEval provides scalable supervision for this task through a constrained and reproducible camera-work schema. Using Qwen3-VL-32B-Instruct enables consistent annotation of a large movie-derived corpus while keeping the label space grounded in standard cinematographic attributes. The evaluator--reviser loop further complements planning by providing an explicit mechanism to identify and refine shots that do not sufficiently satisfy the intended camera work or continuity requirements. Although this iterative procedure introduces additional planning steps, the refinement process is bounded and is designed to improve the reliability of generated multi-shot outputs. The consistent gains across camera-work classification, subject consistency, and aesthetic-quality metrics indicate that structured camera planning transfers effectively to downstream video generation. Future work can further extend this framework through larger-scale human studies, additional video backbones, and broader validation of camera-work annotations across diverse cinematic styles.

\section{Conclusion and Future Work}
We introduce CamPilot, a multi-agent framework for camera-controlled text-to-movie generation. CamPilot separates story planning from cinematographic execution by first constructing scene and shot scripts, then training a Camera Work Planner with GRPO on CamEval to learn camera grammar from real-world professional movies. This design enables structured, controllable camera work that better supports multi-shot continuity and improves overall cinematic quality. Experiments on CamEval show that CamPilot consistently improves camera-work classification and downstream generation quality across multiple planner backbones compared with baselines. In future work, we will extend CamPilot to stronger video backbones and richer evaluation protocols, and explore more adaptive planning and revision strategies to improve long-form coherence, character consistency, and user-controllable cinematic styles.

\section*{Limitations}

While CamPilot demonstrates promising results in improving camera-work planning and controllable text-to-movie generation, we acknowledge several limitations of the current work:\\
$\bullet$ \textbf{\textit{Coverage of Cinematographic Language.}} CamPilot learns camera grammar from a finite set of professional movies and a fixed label schema. As a result, the planner may not fully capture rare or highly stylized cinematographic patterns, such as unconventional lens behavior, complex blocking, or genre-specific shot conventions. In addition, our discrete camera-work labels can underrepresent continuous creative variations in camera movement and composition. Future work can expand coverage by enlarging the source corpus, refining the taxonomy, and exploring discrete-continuous representations that better reflect real cinematography.\\
$\bullet$ \textbf{\textit{Planning Faithfulness and Error Propagation.}} CamPilot relies on a planning-first pipeline, where the shooting script and camera-work plan guide downstream generation. If the planner misinterprets the prompt intent or outputs an implausible plan, the resulting video can deviate from the desired narrative or exhibit inconsistent motion, and such errors can propagate across shots. While we design the planner context to provide scene information, planning faithfulness remains a key bottleneck for multi-shot generation. Future work will investigate stronger plan validation, self-consistency checks, and constrained decoding to reduce invalid or contradictory camera plans before execution.\\
$\bullet$ \textbf{\textit{Generator Dependence and Limited Control Authority.}} Our evaluation fixes the downstream video generator to a single Firefly setting to isolate the effect of camera-work planning. This choice improves experimental control, but it limits conclusions about generalization across generators and about the absolute controllability achievable in different systems. Moreover, even with explicit camera controls, the generator may not perfectly follow the planned camera work due to model limitations or conflicts between content realization and motion constraints. Future work can broaden evaluation to additional generators and study control-aware training or feedback mechanisms that align generation behavior with planned camera trajectories.\\
$\bullet$ \textbf{\textit{Benchmark Scope and Human Preference Alignment.}} CamEval focuses on camera work quality and cinematic engagement, but it does not exhaustively cover all factors that influence user satisfaction, such as storytelling coherence, visual realism, and aesthetic preference. In addition, automatic metrics may not fully reflect human judgments of cinematic quality, especially for subtle transitions or creative shot choices. Future work can extend CamEval with richer human evaluation protocols, preference modeling, and task-specific rubrics that better capture the end-to-end movie creation experience.

\section*{Ethics Statement}
\addcontentsline{toc}{section}{Ethics Statement}
\phantomsection
\label{ethics}
After reviewing the ACL Ethics Policy, we confirm that this work complies with the relevant ethical guidelines. All data used in this study are derived from publicly available sources, and we do not use private or personally identifying information.

\section*{Acknowledgment}
We gratefully acknowledge the support and collaboration from Worcester Polytechnic Institute and Adobe Research. We also thank the reviewers and colleagues for their helpful comments and feedback.

\bibliography{anthology,custom}

\begin{thebibliography}{59}
\expandafter\ifx\csname natexlab\endcsname\relax\def\natexlab#1{#1}\fi

\bibitem[{{Anthropic}(2025)}]{claude_sonnet_45_official}
{Anthropic}. 2025.
\newblock Claude sonnet 4.5.
\newblock \url{https://www.anthropic.com/news/claude-sonnet-4-5}.
\newblock Accessed: 2026-01-06.

\bibitem[{Babu et~al.(2025)Babu, Mahapatra, Rangwani, Soundararajan, and Kulkarni}]{babu2025dynamiceval}
Nithin~C Babu, Aniruddha Mahapatra, Harsh Rangwani, Rajiv Soundararajan, and Kuldeep Kulkarni. 2025.
\newblock Dynamiceval: Rethinking evaluation for dynamic text-to-video synthesis.
\newblock \emph{arXiv preprint arXiv:2510.07441}.

\bibitem[{Bai et~al.(2025)Bai, Cai, Chen, Chen, Chen, Cheng, Deng, Ding, Gao, Ge et~al.}]{bai2025qwen3}
Shuai Bai, Yuxuan Cai, Ruizhe Chen, Keqin Chen, Xionghui Chen, Zesen Cheng, Lianghao Deng, Wei Ding, Chang Gao, Chunjiang Ge, et~al. 2025.
\newblock Qwen3-vl technical report.
\newblock \emph{arXiv preprint arXiv:2511.21631}.

\bibitem[{Bain et~al.(2020)Bain, Nagrani, Brown, and Zisserman}]{bain2020condensed}
Max Bain, Arsha Nagrani, Andrew Brown, and Andrew Zisserman. 2020.
\newblock Condensed movies: Story based retrieval with contextual embeddings.
\newblock In \emph{Asian Conference on Computer Vision}, pages 460--479. Springer.

\bibitem[{Bar-Tal et~al.(2024)Bar-Tal, Chefer, Tov, Herrmann, Paiss, Zada, Ephrat, Hur, Liu, Raj et~al.}]{bar2024lumiere}
Omer Bar-Tal, Hila Chefer, Omer Tov, Charles Herrmann, Roni Paiss, Shiran Zada, Ariel Ephrat, Junhwa Hur, Guanghui Liu, Amit Raj, et~al. 2024.
\newblock Lumiere: A space-time diffusion model for video generation.
\newblock In \emph{SIGGRAPH Asia 2024 Conference Papers}, pages 1--11.

\bibitem[{Blattmann et~al.(2023{\natexlab{a}})Blattmann, Dockhorn, Kulal, Mendelevitch, Kilian, Lorenz, Levi, English, Voleti, Letts et~al.}]{blattmann2023svd}
Andreas Blattmann, Tim Dockhorn, Sumith Kulal, Daniel Mendelevitch, Maciej Kilian, Dominik Lorenz, Yam Levi, Zion English, Vikram Voleti, Adam Letts, et~al. 2023{\natexlab{a}}.
\newblock Stable video diffusion: Scaling latent video diffusion models to large datasets.
\newblock \emph{arXiv preprint arXiv:2311.15127}.

\bibitem[{Blattmann et~al.(2023{\natexlab{b}})Blattmann, Rombach, Ling, Dockhorn, Kim, Fidler, and Kreis}]{blattmann2023align}
Andreas Blattmann, Robin Rombach, Huan Ling, Tim Dockhorn, Seung~Wook Kim, Sanja Fidler, and Karsten Kreis. 2023{\natexlab{b}}.
\newblock Align your latents: High-resolution video synthesis with latent diffusion models.
\newblock In \emph{Proceedings of the IEEE/CVF conference on computer vision and pattern recognition}, pages 22563--22575.

\bibitem[{Brown et~al.(2020)Brown, Mann, Ryder, Subbiah, Kaplan, Dhariwal, Neelakantan, Shyam, Sastry, Askell et~al.}]{brown2020gpt3}
Tom Brown, Benjamin Mann, Nick Ryder, Melanie Subbiah, Jared~D Kaplan, Prafulla Dhariwal, Arvind Neelakantan, Pranav Shyam, Girish Sastry, Amanda Askell, et~al. 2020.
\newblock Language models are few-shot learners.
\newblock \emph{Advances in neural information processing systems}, 33:1877--1901.

\bibitem[{Cen et~al.(2025)Cen, Meng, Hu, Liu, and Wu}]{cen2025pde}
Mengqi Cen, Xuejing Meng, X~Joan Hu, Juxin Liu, and Jianhong Wu. 2025.
\newblock Pde-based bayesian hierarchical modeling for event spread, with application to covid-19 infection.
\newblock \emph{arXiv preprint arXiv:2509.13174}.

\bibitem[{Chen et~al.(2025)Chen, Guo, Shi, Song, and Zhang}]{chen2025t2vworldbench}
Yubin Chen, Xuyang Guo, Zhenmei Shi, Zhao Song, and Jiahao Zhang. 2025.
\newblock T2vworldbench: A benchmark for evaluating world knowledge in text-to-video generation.
\newblock \emph{arXiv preprint arXiv:2507.18107}.

\bibitem[{Courant et~al.(2024)Courant, Dufour, Wang, Christie, and Kalogeiton}]{courant2024exceptional}
Robin Courant, Nicolas Dufour, Xi~Wang, Marc Christie, and Vicky Kalogeiton. 2024.
\newblock Et the exceptional trajectories: Text-to-camera-trajectory generation with character awareness.
\newblock In \emph{European Conference on Computer Vision}, pages 464--480. Springer.

\bibitem[{{DeepSeek AI}(2025)}]{deepseek_r1_modelcard}
{DeepSeek AI}. 2025.
\newblock deepseek-ai/deepseek-r1 model card.
\newblock \url{https://huggingface.co/deepseek-ai/DeepSeek-R1}.
\newblock Accessed: 2026-01-06.

\bibitem[{Feng et~al.(2024)Feng, Liu, Tu, Qi, Sun, Ma, Zhao, Zhou, and He}]{feng2024i2vcontrol}
Wanquan Feng, Jiawei Liu, Pengqi Tu, Tianhao Qi, Mingzhen Sun, Tianxiang Ma, Songtao Zhao, Siyu Zhou, and Qian He. 2024.
\newblock I2vcontrol-camera: Precise video camera control with adjustable motion strength.
\newblock \emph{arXiv preprint arXiv:2411.06525}.

\bibitem[{Geng et~al.(2025)Geng, Herrmann, Hur, Cole, Zhang, Pfaff, Lopez-Guevara, Aytar, Rubinstein, Sun et~al.}]{geng2025motion}
Daniel Geng, Charles Herrmann, Junhwa Hur, Forrester Cole, Serena Zhang, Tobias Pfaff, Tatiana Lopez-Guevara, Yusuf Aytar, Michael Rubinstein, Chen Sun, et~al. 2025.
\newblock Motion prompting: Controlling video generation with motion trajectories.
\newblock In \emph{Proceedings of the Computer Vision and Pattern Recognition Conference}, pages 1--12.

\bibitem[{{Google}(2025)}]{gemma3_12b_it_modelcard}
{Google}. 2025.
\newblock google/gemma-3-12b-it model card.
\newblock \url{https://huggingface.co/google/gemma-3-12b-it}.
\newblock Accessed: 2026-01-06.

\bibitem[{He et~al.(2024)He, Xu, Guo, Wetzstein, Dai, Li, and Yang}]{he2024cameractrl}
Hao He, Yinghao Xu, Yuwei Guo, Gordon Wetzstein, Bo~Dai, Hongsheng Li, and Ceyuan Yang. 2024.
\newblock Cameractrl: Enabling camera control for text-to-video generation.
\newblock \emph{arXiv preprint arXiv:2404.02101}.

\bibitem[{Ho et~al.(2022)Ho, Salimans, Gritsenko, Chan, Norouzi, and Fleet}]{ho2022video}
Jonathan Ho, Tim Salimans, Alexey Gritsenko, William Chan, Mohammad Norouzi, and David~J Fleet. 2022.
\newblock Video diffusion models.
\newblock \emph{Advances in neural information processing systems}, 35:8633--8646.

\bibitem[{Hong et~al.(2023)Hong, Zhuge, Chen, Zheng, Cheng, Wang, Zhang, Wang, Yau, Lin et~al.}]{hong2023metagpt}
Sirui Hong, Mingchen Zhuge, Jonathan Chen, Xiawu Zheng, Yuheng Cheng, Jinlin Wang, Ceyao Zhang, Zili Wang, Steven Ka~Shing Yau, Zijuan Lin, et~al. 2023.
\newblock Metagpt: Meta programming for a multi-agent collaborative framework.
\newblock In \emph{The Twelfth International Conference on Learning Representations}.

\bibitem[{Hu et~al.(2024)Hu, Jiang, Chen, Han, Liao, Chang, and Liang}]{hu2024storyagent}
Panwen Hu, Jin Jiang, Jianqi Chen, Mingfei Han, Shengcai Liao, Xiaojun Chang, and Xiaodan Liang. 2024.
\newblock Storyagent: Customized storytelling video generation via multi-agent collaboration.
\newblock \emph{arXiv preprint arXiv:2411.04925}.

\bibitem[{Huang et~al.(2024)Huang, He, Yu, Zhang, Si, Jiang, Zhang, Wu, Jin, Chanpaisit et~al.}]{huang2024vbench}
Ziqi Huang, Yinan He, Jiashuo Yu, Fan Zhang, Chenyang Si, Yuming Jiang, Yuanhan Zhang, Tianxing Wu, Qingyang Jin, Nattapol Chanpaisit, et~al. 2024.
\newblock Vbench: Comprehensive benchmark suite for video generative models.
\newblock In \emph{Proceedings of the IEEE/CVF Conference on Computer Vision and Pattern Recognition}, pages 21807--21818.

\bibitem[{Hui et~al.(2024)Hui, Yang, Cui, Yang, Liu, Zhang, Liu, Zhang, Yu, Lu et~al.}]{hui2024qwen2}
Binyuan Hui, Jian Yang, Zeyu Cui, Jiaxi Yang, Dayiheng Liu, Lei Zhang, Tianyu Liu, Jiajun Zhang, Bowen Yu, Keming Lu, et~al. 2024.
\newblock Qwen2. 5-coder technical report.
\newblock \emph{arXiv preprint arXiv:2409.12186}.

\bibitem[{Kondratyuk et~al.(2023)Kondratyuk, Yu, Gu, Lezama, Huang, Schindler, Hornung, Birodkar, Yan, Chiu et~al.}]{kondratyuk2023videopoet}
Dan Kondratyuk, Lijun Yu, Xiuye Gu, Jos{\'e} Lezama, Jonathan Huang, Grant Schindler, Rachel Hornung, Vighnesh Birodkar, Jimmy Yan, Ming-Chang Chiu, et~al. 2023.
\newblock Videopoet: A large language model for zero-shot video generation.
\newblock \emph{arXiv preprint arXiv:2312.14125}.

\bibitem[{Kuang et~al.(2024)Kuang, Cai, He, Xu, Li, Guibas, and Wetzstein}]{kuang2024collaborative}
Zhengfei Kuang, Shengqu Cai, Hao He, Yinghao Xu, Hongsheng Li, Leonidas~J Guibas, and Gordon Wetzstein. 2024.
\newblock Collaborative video diffusion: Consistent multi-video generation with camera control.
\newblock \emph{Advances in Neural Information Processing Systems}, 37:16240--16271.

\bibitem[{Li et~al.(2023)Li, Hammoud, Itani, Khizbullin, and Ghanem}]{li2023camel}
Guohao Li, Hasan Hammoud, Hani Itani, Dmitrii Khizbullin, and Bernard Ghanem. 2023.
\newblock Camel: Communicative agents for" mind" exploration of large language model society.
\newblock \emph{Advances in Neural Information Processing Systems}, 36:51991--52008.

\bibitem[{Li et~al.(2024{\natexlab{a}})Li, Wu, Yang, Qu, Zhang, Chen, Li, Mu, Hu, Fang et~al.}]{li2024can}
Xiaozhe Li, Kai Wu, Siyi Yang, YiZhan Qu, Guohua Zhang, Zhiyu Chen, Jiayao Li, Jiangchuan Mu, Xiaobin Hu, Wen Fang, et~al. 2024{\natexlab{a}}.
\newblock Can video generation replace cinematographers? research on the cinematic language of generated video.
\newblock \emph{arXiv preprint arXiv:2412.12223}.

\bibitem[{Li et~al.(2024{\natexlab{b}})Li, Shi, Hu, Wang, Zhu, Xu, Zhao, and Zhang}]{li2024animdirector}
Yunxin Li, Haoyuan Shi, Baotian Hu, Longyue Wang, Jiashun Zhu, Jinyi Xu, Zhen Zhao, and Min Zhang. 2024{\natexlab{b}}.
\newblock Anim-director: A large multimodal model powered agent for controllable animation video generation.
\newblock In \emph{SIGGRAPH Asia 2024 Conference Papers}, pages 1--11.

\bibitem[{Lin et~al.(2023)Lin, Zala, Cho, and Bansal}]{lin2023videodirectorgpt}
Han Lin, Abhay Zala, Jaemin Cho, and Mohit Bansal. 2023.
\newblock Videodirectorgpt: Consistent multi-scene video generation via llm-guided planning.
\newblock \emph{arXiv preprint arXiv:2309.15091}.

\bibitem[{Ling et~al.(2025)Ling, Zhu, Wu, Li, Feng, Yang, Hao, Zhu, Wu, and Chu}]{ling2025vmbench}
Xinran Ling, Chen Zhu, Meiqi Wu, Hangyu Li, Xiaokun Feng, Cundian Yang, Aiming Hao, Jiashu Zhu, Jiahong Wu, and Xiangxiang Chu. 2025.
\newblock Vmbench: A benchmark for perception-aligned video motion generation.
\newblock In \emph{Proceedings of the IEEE/CVF International Conference on Computer Vision}, pages 13087--13098.

\bibitem[{Long et~al.(2024)Long, Qiu, Yao, and Mei}]{long2024videostudio}
Fuchen Long, Zhaofan Qiu, Ting Yao, and Tao Mei. 2024.
\newblock Videostudio: Generating consistent-content and multi-scene videos.
\newblock In \emph{European Conference on Computer Vision}, pages 468--485. Springer.

\bibitem[{{Meta}(2024)}]{llama31_8b_instruct_modelcard}
{Meta}. 2024.
\newblock meta-llama/llama-3.1-8b-instruct model card.
\newblock \url{https://huggingface.co/meta-llama/Llama-3.1-8B-Instruct}.
\newblock Accessed: 2026-01-06.

\bibitem[{Nakano et~al.(2021)Nakano, Hilton, Balaji, Wu, Ouyang, Kim, Hesse, Jain, Kosaraju, Saunders et~al.}]{nakano2021webgpt}
Reiichiro Nakano, Jacob Hilton, Suchir Balaji, Jeff Wu, Long Ouyang, Christina Kim, Christopher Hesse, Shantanu Jain, Vineet Kosaraju, William Saunders, et~al. 2021.
\newblock Webgpt: Browser-assisted question-answering with human feedback.
\newblock \emph{arXiv preprint arXiv:2112.09332}.

\bibitem[{{OpenAI}(2025)}]{gpt5_official}
{OpenAI}. 2025.
\newblock Introducing gpt-5.
\newblock \url{https://openai.com/index/introducing-gpt-5/}.
\newblock Accessed: 2026-01-06.

\bibitem[{Park et~al.(2023)Park, O'Brien, Cai, Morris, Liang, and Bernstein}]{park2023generativeagents}
Joon~Sung Park, Joseph O'Brien, Carrie~Jun Cai, Meredith~Ringel Morris, Percy Liang, and Michael~S Bernstein. 2023.
\newblock Generative agents: Interactive simulacra of human behavior.
\newblock In \emph{Proceedings of the 36th annual acm symposium on user interface software and technology}, pages 1--22.

\bibitem[{Paszke et~al.(2019)Paszke, Gross, Massa, Lerer, Bradbury, Chanan, Killeen, Lin, Gimelshein, Antiga et~al.}]{paszke2019pytorch}
Adam Paszke, Sam Gross, Francisco Massa, Adam Lerer, James Bradbury, Gregory Chanan, Trevor Killeen, Zeming Lin, Natalia Gimelshein, Luca Antiga, et~al. 2019.
\newblock Pytorch: An imperative style, high-performance deep learning library.
\newblock \emph{Advances in neural information processing systems}, 32.

\bibitem[{Qian et~al.(2024)Qian, Liu, Liu, Chen, Dang, Li, Yang, Chen, Su, Cong et~al.}]{qian2023chatdev}
Chen Qian, Wei Liu, Hongzhang Liu, Nuo Chen, Yufan Dang, Jiahao Li, Cheng Yang, Weize Chen, Yusheng Su, Xin Cong, et~al. 2024.
\newblock Chatdev: Communicative agents for software development.
\newblock In \emph{Proceedings of the 62nd Annual Meeting of the Association for Computational Linguistics (Volume 1: Long Papers)}, pages 15174--15186.

\bibitem[{Qin et~al.(2023)Qin, Liang, Ye, Zhu, Yan, Lu, Lin, Cong, Tang, Qian et~al.}]{qin2023toollm}
Yujia Qin, Shihao Liang, Yining Ye, Kunlun Zhu, Lan Yan, Yaxi Lu, Yankai Lin, Xin Cong, Xiangru Tang, Bill Qian, et~al. 2023.
\newblock Toolllm: Facilitating large language models to master 16000+ real-world apis.
\newblock \emph{arXiv preprint arXiv:2307.16789}.

\bibitem[{Radford et~al.(2021)Radford, Kim, Hallacy, Ramesh, Goh, Agarwal, Sastry, Askell, Mishkin, Clark et~al.}]{radford2021learning}
Alec Radford, Jong~Wook Kim, Chris Hallacy, Aditya Ramesh, Gabriel Goh, Sandhini Agarwal, Girish Sastry, Amanda Askell, Pamela Mishkin, Jack Clark, et~al. 2021.
\newblock Learning transferable visual models from natural language supervision.
\newblock In \emph{International conference on machine learning}, pages 8748--8763. PmLR.

\bibitem[{Salimans et~al.(2016)Salimans, Goodfellow, Zaremba, Cheung, Radford, and Chen}]{salimans2016improved}
Tim Salimans, Ian Goodfellow, Wojciech Zaremba, Vicki Cheung, Alec Radford, and Xi~Chen. 2016.
\newblock Improved techniques for training gans.
\newblock \emph{Advances in neural information processing systems}, 29.

\bibitem[{Schick et~al.(2023)Schick, Dwivedi-Yu, Dess{\`\i}, Raileanu, Lomeli, Hambro, Zettlemoyer, Cancedda, and Scialom}]{schick2023toolformer}
Timo Schick, Jane Dwivedi-Yu, Roberto Dess{\`\i}, Roberta Raileanu, Maria Lomeli, Eric Hambro, Luke Zettlemoyer, Nicola Cancedda, and Thomas Scialom. 2023.
\newblock Toolformer: Language models can teach themselves to use tools.
\newblock \emph{Advances in Neural Information Processing Systems}, 36:68539--68551.

\bibitem[{Shao et~al.(2024)Shao, Wang, Zhu, Xu, Song, Bi, Zhang, Zhang, Li, Wu et~al.}]{shao2024deepseekmath}
Zhihong Shao, Peiyi Wang, Qihao Zhu, Runxin Xu, Junxiao Song, Xiao Bi, Haowei Zhang, Mingchuan Zhang, YK~Li, Yang Wu, et~al. 2024.
\newblock Deepseekmath: Pushing the limits of mathematical reasoning in open language models.
\newblock \emph{arXiv preprint arXiv:2402.03300}.

\bibitem[{Shen et~al.(2023)Shen, Song, Tan, Li, Lu, and Zhuang}]{shen2023hugginggpt}
Yongliang Shen, Kaitao Song, Xu~Tan, Dongsheng Li, Weiming Lu, and Yueting Zhuang. 2023.
\newblock Hugginggpt: Solving ai tasks with chatgpt and its friends in hugging face.
\newblock \emph{Advances in Neural Information Processing Systems}, 36:38154--38180.

\bibitem[{Shinn et~al.(2023)Shinn, Cassano, Gopinath, Narasimhan, and Yao}]{shinn2023reflexion}
Noah Shinn, Federico Cassano, Ashwin Gopinath, Karthik Narasimhan, and Shunyu Yao. 2023.
\newblock Reflexion: Language agents with verbal reinforcement learning.
\newblock \emph{Advances in Neural Information Processing Systems}, 36:8634--8652.

\bibitem[{Singer et~al.(2022)Singer, Polyak, Hayes, Yin, An, Zhang, Hu, Yang, Ashual, Gafni et~al.}]{singer2022makeavideo}
Uriel Singer, Adam Polyak, Thomas Hayes, Xi~Yin, Jie An, Songyang Zhang, Qiyuan Hu, Harry Yang, Oron Ashual, Oran Gafni, et~al. 2022.
\newblock Make-a-video: Text-to-video generation without text-video data.
\newblock \emph{arXiv preprint arXiv:2209.14792}.

\bibitem[{von Werra et~al.(2022)von Werra, Belkada, Tunstall, Beeching, Thrush, Lambert, Huang, Rasul, and Gallou\"{e}dec}]{vonwerra2022trl}
Leandro von Werra, Younes Belkada, Luke Tunstall, Edward Beeching, Thomas Thrush, Nathan Lambert, Shixiang Huang, Kashif Rasul, and Quentin Gallou\"{e}dec. 2022.
\newblock Trl: Transformer reinforcement learning library.
\newblock \url{https://github.com/huggingface/trl}.
\newblock Accessed: 2025-11.

\bibitem[{Wang et~al.(2023{\natexlab{a}})Wang, Xie, Jiang, Mandlekar, Xiao, Zhu, Fan, and Anandkumar}]{wang2023voyager}
Guanzhi Wang, Yuqi Xie, Yunfan Jiang, Ajay Mandlekar, Chaowei Xiao, Yuke Zhu, Linxi Fan, and Anima Anandkumar. 2023{\natexlab{a}}.
\newblock Voyager: An open-ended embodied agent with large language models.
\newblock \emph{arXiv preprint arXiv:2305.16291}.

\bibitem[{Wang et~al.(2023{\natexlab{b}})Wang, Yuan, Chen, Zhang, Wang, and Zhang}]{wang2023modelscope}
Jiuniu Wang, Hangjie Yuan, Dayou Chen, Yingya Zhang, Xiang Wang, and Shiwei Zhang. 2023{\natexlab{b}}.
\newblock Modelscope text-to-video technical report.
\newblock \emph{arXiv preprint arXiv:2308.06571}.

\bibitem[{Wang et~al.(2025)Wang, Chen, Ma, Zhou, Huang, Wang, Yang, He, Yu, Yang et~al.}]{wang2023lavie}
Yaohui Wang, Xinyuan Chen, Xin Ma, Shangchen Zhou, Ziqi Huang, Yi~Wang, Ceyuan Yang, Yinan He, Jiashuo Yu, Peiqing Yang, et~al. 2025.
\newblock Lavie: High-quality video generation with cascaded latent diffusion models.
\newblock \emph{International Journal of Computer Vision}, 133(5):3059--3078.

\bibitem[{Wei et~al.(2022)Wei, Wang, Schuurmans, Bosma, Xia, Chi, Le, Zhou et~al.}]{wei2022cot}
Jason Wei, Xuezhi Wang, Dale Schuurmans, Maarten Bosma, Fei Xia, Ed~Chi, Quoc~V Le, Denny Zhou, et~al. 2022.
\newblock Chain-of-thought prompting elicits reasoning in large language models.
\newblock \emph{Advances in neural information processing systems}, 35:24824--24837.

\bibitem[{Wolf et~al.(2020)Wolf, Debut, Sanh, Chaumond, Delangue, Moi, Cistac, Rault, Louf, Funtowicz et~al.}]{wolf2020transformers}
Thomas Wolf, Lysandre Debut, Victor Sanh, Julien Chaumond, Clement Delangue, Anthony Moi, Pierric Cistac, Tim Rault, Remi Louf, Morgan Funtowicz, et~al. 2020.
\newblock Transformers: State-of-the-art natural language processing.
\newblock In \emph{Proceedings of the 2020 conference on empirical methods in natural language processing: system demonstrations}, pages 38--45.

\bibitem[{Wu et~al.(2025{\natexlab{a}})Wu, Zhu, and Shou}]{wu2025automated}
Weijia Wu, Zeyu Zhu, and Mike~Zheng Shou. 2025{\natexlab{a}}.
\newblock Automated movie generation via multi-agent cot planning.
\newblock \emph{arXiv preprint arXiv:2503.07314}.

\bibitem[{Wu et~al.(2024)Wu, Wang, Gumusel, and Liu}]{wu2024knowledge}
Yang Wu, Chenghao Wang, Ece Gumusel, and Xiaozhong Liu. 2024.
\newblock Knowledge-infused legal wisdom: Navigating llm consultation through the lens of diagnostics and positive-unlabeled reinforcement learning.
\newblock In \emph{Findings of the Association for Computational Linguistics: ACL 2024}, pages 15542--15555.

\bibitem[{Wu et~al.(2025{\natexlab{b}})Wu, Yao, Zhang, Shi, Jiang, Li, and Liu}]{wu2025teaching}
Yang Wu, Rujing Yao, Tong Zhang, Yufei Shi, Zhuoren Jiang, Zhushan Li, and Xiaozhong Liu. 2025{\natexlab{b}}.
\newblock Teaching according to students' aptitude: Personalized mathematics tutoring via persona-, memory-, and forgetting-aware llms.
\newblock \emph{arXiv preprint arXiv:2511.15163}.

\bibitem[{Xie et~al.(2024)Xie, Tang, Tan, Klein, Bissyand, and Ezzini}]{xie2024dreamfactory}
Zhifei Xie, Daniel Tang, Dingwei Tan, Jacques Klein, Tegawend~F Bissyand, and Saad Ezzini. 2024.
\newblock Dreamfactory: Pioneering multi-scene long video generation with a multi-agent framework.
\newblock \emph{arXiv preprint arXiv:2408.11788}.

\bibitem[{Xing et~al.(2025)Xing, Mai, Ham, Huang, Mahapatra, Fu, Wong, and Liu}]{xing2025motioncanvas}
Jinbo Xing, Long Mai, Cusuh Ham, Jiahui Huang, Aniruddha Mahapatra, Chi-Wing Fu, Tien-Tsin Wong, and Feng Liu. 2025.
\newblock Motioncanvas: Cinematic shot design with controllable image-to-video generation.
\newblock In \emph{Proceedings of the Special Interest Group on Computer Graphics and Interactive Techniques Conference Conference Papers}, pages 1--11.

\bibitem[{Xu et~al.(2025)Xu, Wang, Wang, Li, Shi, Yang, Wang, Hu, Yu, and Zhang}]{xu2025filmagent}
Zhenran Xu, Longyue Wang, Jifang Wang, Zhouyi Li, Senbao Shi, Xue Yang, Yiyu Wang, Baotian Hu, Jun Yu, and Min Zhang. 2025.
\newblock Filmagent: A multi-agent framework for end-to-end film automation in virtual 3d spaces.
\newblock \emph{arXiv preprint arXiv:2501.12909}.

\bibitem[{Yan et~al.(2021)Yan, Zhang, Abbeel, and Srinivas}]{yan2021videogpt}
Wilson Yan, Yunzhi Zhang, Pieter Abbeel, and Aravind Srinivas. 2021.
\newblock Videogpt: Video generation using vq-vae and transformers.
\newblock \emph{arXiv preprint arXiv:2104.10157}.

\bibitem[{Yao et~al.(2025)Yao, Wu, Wang, Xiong, Wang, and Liu}]{yao2025elevating}
Rujing Yao, Yang Wu, Chenghao Wang, Jingwei Xiong, Fang Wang, and Xiaozhong Liu. 2025.
\newblock Elevating legal llm responses: harnessing trainable logical structures and semantic knowledge with legal reasoning.
\newblock In \emph{Proceedings of the 2025 Conference of the Nations of the Americas Chapter of the Association for Computational Linguistics: Human Language Technologies (Volume 1: Long Papers)}, pages 5630--5642.

\bibitem[{Yao et~al.(2023)Yao, Yu, Zhao, Shafran, Griffiths, Cao, and Narasimhan}]{yao2023tot}
Shunyu Yao, Dian Yu, Jeffrey Zhao, Izhak Shafran, Tom Griffiths, Yuan Cao, and Karthik Narasimhan. 2023.
\newblock Tree of thoughts: Deliberate problem solving with large language models.
\newblock \emph{Advances in neural information processing systems}, 36:11809--11822.

\bibitem[{Yuan et~al.(2024)Yuan, Liu, Cao, Sun, Jia, Chen, Li, Lin, Yuan, He et~al.}]{mora2024}
Zhengqing Yuan, Yixin Liu, Yihan Cao, Weixiang Sun, Haolong Jia, Ruoxi Chen, Zhaoxu Li, Bin Lin, Li~Yuan, Lifang He, et~al. 2024.
\newblock Mora: Enabling generalist video generation via a multi-agent framework.
\newblock \emph{arXiv preprint arXiv:2403.13248}.

\end{thebibliography}
\bibliographystyle{acl_natbib}

\nocite{}
\clearpage 

\appendix

\section{Performance Comparison on Subtasks}

Table~\ref{tab:subtasks_result} compares subtask-level camera work classification across camera angle, shot size, and multiple motion attributes. CamPilot consistently achieves the best Macro-ACC on all subtasks, for example improving Camera Angle Macro-ACC to 92.7 (vs.\ 84.0 for MovieAgent and 80.4 for DreamFactory) and Shot Size Macro-ACC to 83.7 (vs.\ 78.3 for Standard). The gains are especially clear on motion-related subtasks: on Camera Motion (Translation), CamPilot reaches 61.2 Macro-ACC with 31.0 Macro-Prec, 30.4 Macro-Rec, and 28.3 Macro-F1, outperforming the strongest baseline (MovieAgent) at 51.8/25.3/24.4/24.2. Similarly, on Camera Motion (Rotation), CamPilot achieves 86.1 Macro-ACC and leads Macro-Prec/Macro-Rec/Macro-F1 at 25.5/28.9/26.2, while MovieAgent attains 70.4 Macro-ACC. On Focal Length, CamPilot further raises Macro-ACC to 92.2 and recall to 41.0, compared to 87.1 Macro-ACC and 31.7 recall for MovieAgent. Overall, these results show that CamPilot improves both correctness and robustness across diverse subtasks, with the largest margins appearing on motion translation, rotation, and focal length where cross-shot camera reasoning is most needed.

\begin{table*}[t]
\centering
\small
\renewcommand\arraystretch{0.97}

\resizebox{0.8\textwidth}{!}{
\begin{tabular}{llcccc} 
\toprule
Task & Method & Macro-ACC & Macro-Prec & Macro-Rec & Macro-F1  \\
\midrule

\multirow{5}{*}{Camera Angle} 
& Standard         & 66.9 (0.4) & 2.8 (0.2) & 2.9 (0.2) & 2.5 (0.2) \\
& Vanilla-SFT      & \underline{86.0 (0.3)} & 11.3 (0.5) & 11.4 (0.5) & 11.3 (0.5) \\
& DreamFactory     & 80.4 (0.4) & 34.3 (0.7) & \textbf{29.9 (0.6)} & \textbf{25.0 (0.6)} \\
& MovieAgent       & 84.0 (0.3) & \underline{34.8 (0.7)} & \underline{28.5 (0.6)} & \underline{24.9 (0.6)} \\
& CamPilot (Ours)  & \textbf{92.7 (0.2)} & \textbf{41.3 (0.6)} & 22.0 (0.6) & 23.0 (0.6) \\
\midrule

\multirow{5}{*}{Shot Size} 
& Standard         & \underline{78.3 (0.4)} & 9.9 (0.5) & 7.4 (0.4) & 7.7 (0.4) \\
& Vanilla-SFT      & 75.8 (0.4) & 11.7 (0.5) & 11.3 (0.5) & 11.4 (0.5) \\
& DreamFactory     & 66.2 (0.5) & \underline{37.0 (0.7)} & \textbf{41.0 (0.8)} & \underline{33.5 (0.7)} \\
& MovieAgent       & 76.1 (0.4) & 31.2 (0.7) & 25.6 (0.7) & 26.6 (0.7) \\
& CamPilot (Ours)  & \textbf{83.7 (0.3)} & \textbf{37.1 (0.7)} & \underline{35.7 (0.7)} & \textbf{34.8 (0.7)} \\ 
\midrule

\multirow{5}{*}{\begin{tabular}[t]{@{}l@{}}Camera Motion\\(Type)\end{tabular}}
& Standard         & 42.0 (0.5) & 1.8 (0.2) & 1.5 (0.2) & 1.4 (0.2) \\
& Vanilla-SFT      & 43.5 (0.5) & 3.4 (0.3) & 3.1 (0.3) & 3.2 (0.3) \\
& DreamFactory     & 38.3 (0.5) & \underline{8.5 (0.5)} & \textbf{9.3 (0.5)} & \underline{7.7 (0.5)} \\
& MovieAgent       & \underline{49.9 (0.4)} & 7.6 (0.5) & 6.0 (0.5) & 5.9 (0.5) \\
& CamPilot (Ours)  & \textbf{59.4 (0.3)} & \textbf{11.1 (0.5)} & \underline{8.2 (0.5)} & \textbf{8.7 (0.5)} \\
\midrule

\multirow{5}{*}{\begin{tabular}[t]{@{}l@{}}Camera Motion\\(Translation)\end{tabular}}
& Standard         & 48.3 (0.5) & 13.9 (0.6) & 12.8 (0.6) & 12.3 (0.6) \\
& Vanilla-SFT      & 45.7 (0.5) & 17.4 (0.6) & 17.0 (0.6) & 17.1 (0.6) \\
& DreamFactory     & 48.5 (0.5) & 22.5 (0.7) & 21.6 (0.7) & 21.3 (0.7) \\
& MovieAgent       & \underline{51.8 (0.4)} & \underline{25.3 (0.7)} & \underline{24.4 (0.7)} & \underline{24.2 (0.7)} \\
& CamPilot (Ours)  & \textbf{61.2 (0.3)} & \textbf{31.0 (0.7)} & \textbf{30.4 (0.7)} & \textbf{28.3 (0.7)} \\
\midrule

\multirow{5}{*}{\begin{tabular}[t]{@{}l@{}}Camera Motion\\(Rotation)\end{tabular}}
& Standard         & \underline{80.0 (0.4)} & 11.2 (0.6) & 10.8 (0.6) & 10.8 (0.6) \\
& Vanilla-SFT      & 77.6 (0.4) & 15.6 (0.6) & 15.5 (0.6) & 15.5 (0.6) \\
& DreamFactory     & 65.7 (0.5) & \underline{23.4 (0.7)} & \underline{26.3 (0.7)} & \underline{23.4 (0.7)} \\
& MovieAgent       & 70.4 (0.5) & 17.2 (0.6) & 20.0 (0.6) & 18.5 (0.6) \\
& CamPilot (Ours)  & \textbf{86.1 (0.3)} & \textbf{25.5 (0.7)} & \textbf{28.9 (0.7)} & \textbf{26.2 (0.7)}\\
\midrule

\multirow{5}{*}{\begin{tabular}[t]{@{}l@{}}Camera Motion\\(Speed)\end{tabular}}
& Standard         & 49.6 (0.5) & 12.3 (0.6) & 15.6 (0.6) & 11.3 (0.6) \\
& Vanilla-SFT      & 53.1 (0.5) & 25.5 (0.7) & 24.6 (0.7) & 24.9 (0.7) \\
& DreamFactory     & 52.7 (0.5) & 33.4 (0.7) & \textbf{42.6 (0.8)} & 34.2 (0.7) \\
& MovieAgent       & \underline{58.2 (0.4)} & \underline{34.5 (0.7)} & \underline{39.8 (0.8)} & \underline{34.4 (0.7)} \\
& CamPilot (Ours)  & \textbf{63.7 (0.3)} & \textbf{36.5 (0.7)} & 35.4 (0.7) & \textbf{35.2 (0.7)} \\
\midrule

\multirow{5}{*}{\begin{tabular}[t]{@{}l@{}}Camera Motion\\(Focal Length)\end{tabular}}
& Standard         & 74.4 (0.5) & 11.4 (0.6) & 12.7 (0.6) & 10.9 (0.6) \\
& Vanilla-SFT      & 86.1 (0.4) & 23.6 (0.7) & 23.7 (0.7) & 23.6 (0.7) \\
& DreamFactory     & 57.1 (0.6) & 23.1 (0.7) & 25.0 (0.7) & 24.0 (0.7) \\
& MovieAgent       & \underline{87.1 (0.4)} & \textbf{32.1 (0.7)} & \underline{31.7 (0.7)} & \textbf{31.6 (0.7)} \\
& CamPilot (Ours)  & \textbf{92.2 (0.3)} & \underline{28.8 (0.7)} & \textbf{41.0 (0.8)} & \underline{26.2 (0.7)} \\
\bottomrule
\end{tabular}
}
\caption{Sub-task performance (\%) comparison of our method against baselines for multi-shot video generation. The best result is highlighted in \textbf{bold}, and the second-best is \underline{underlined}. \textbf{Average} represents the average scores across all tasks.}
\label{tab:subtasks_result}
\end{table*}

\section{Prompt}
In this section, we present the prompt used for shooting script generation. The prompt frames the model as a professional cinematographer and camera operator for camera-work planning in text-to-movie generation. Given a sequence of time-ordered sampled frames from a short clip, the model is instructed to produce a structured and reproducible annotation that (i) describes the shot content at a high level using only observable visual evidence, (ii) predicts camera-work attributes under a fixed label schema, and (iii) provides a brief intent-oriented rationale grounded in what is visible. The prompt further specifies the annotation scope and temporal assumption: multiple frames are treated as consecutive moments within the same shot unless a cut is evident, in which case the model annotates the dominant shot and records a short cut note. To improve robustness and reproducibility, we include reliability constraints that encourage conservative judgments for ambiguous cases, enforce internal consistency across fields, and avoid subjective or speculative descriptions.

\begin{prompt}{Shooting Script Generation}
\textless\textbar im\_start\textbar\textgreater system\\
You are a professional cinematographer and camera operator for camera-work planning in text-to-movie generation. You will receive a sequence of time-ordered frames from a short clip, and you must produce a structured, reproducible annotation that captures cinematic shot design and camera behavior.

Your responsibilities:\\
- Describe the shot content at a high level using only observable visual evidence.\\
- Infer camera-work attributes using standard cinematography language and a fixed label schema.\\
- Provide a brief intent-oriented rationale grounded in what is visible.\\
- Output strictly valid JSON with the required fields and no extra text.\\

Annotation scope:\\
- Focus on camera-related attributes: shot size, shot angle, camera motion, focal behavior, motion speed, and motion intensity.\\
- If multiple frames are provided, treat them as consecutive moments from the same shot unless a cut is evident.\\
- If a cut is evident, annotate the dominant shot and record a short cut note.\\

Safety and privacy constraints:\\
- Do not infer or reveal identities, personal attributes, or private information.\\
- Do not name real persons, brands, or copyrighted titles.\\
- Do not speculate about unseen events, off-screen objects, or backstory.\\

Reliability guidelines:\\
- Use conservative judgments. If an attribute is ambiguous, select the closest neutral label (e.g., eye-level, medium, static, none, fixed focal length).\\
- Keep the description concrete and technical. Avoid subjective praise or vague cinematic adjectives.\\
- Ensure internal consistency across fields (e.g., a static shot should not have fast translation).\\

Label schema (use exactly one option per field):\\
- camera\_angle: \{eye-level, high-angle, low-angle, overhead, dutch-angle\}\\
- shot\_size: \{extreme close-up, close-up, medium, medium-long, long, extreme long\}\\
- camera\_motion\_type: \{static, pan, tilt, zoom, dolly, tracking, handheld\}\\
- focal\_behavior: \{fixed focal length, zoom-in, zoom-out\}\\
- motion\_speed: \{slow, medium, fast\}\\
- rotation\_intensity: \{none, subtle, strong\}\\
- translation\_intensity: \{none, subtle, strong\}\\

Output format (JSON only):\\
\textless\textbar im\_end\textbar\textgreater\\

\textless\textbar im\_start\textbar\textgreater user\\
Visual input: \textless sampled frames from a clip\textgreater\\
Task: Create one JSON annotation following the schema above.\\
\textless\textbar im\_end\textbar\textgreater
\end{prompt}

\end{document}